\documentclass{article}

\usepackage{microtype}
\usepackage{graphicx}
\usepackage{subfigure}
\usepackage{booktabs} 

\usepackage{hyperref}

\usepackage[accepted]{icml2024}

\usepackage{amsmath}
\usepackage{amssymb}
\usepackage{mathtools}
\usepackage{amsthm}

\usepackage[capitalize,noabbrev]{cleveref}

\theoremstyle{plain}

\theoremstyle{definition}

\theoremstyle{remark}

\usepackage[textsize=tiny]{todonotes}

\icmltitlerunning{Levels of AGI}

\begin{document}

\twocolumn[
\icmltitle{Position: Levels of AGI for
          Operationalizing Progress on the Path to AGI}



\icmlsetsymbol{equal}{*}

\begin{icmlauthorlist}
\icmlauthor{Meredith Ringel Morris}{comp1}
\icmlauthor{Jascha Sohl-Dickstein}{comp2}
\icmlauthor{Noah Fiedel}{comp2}
\icmlauthor{Tris Wartkentin}{comp2}
\icmlauthor{Allan Dafoe}{comp3}
\icmlauthor{Aleksandra Faust}{comp2}
\icmlauthor{Clement Farbaret}{comp3}
\icmlauthor{Shane Legg}{comp3}
\end{icmlauthorlist}

\icmlaffiliation{comp1}{Google DeepMind, Seattle, WA, USA}
\icmlaffiliation{comp2}{Google DeepMind, Mountain View, CA, USA}
\icmlaffiliation{comp3}{Google DeepMind, London, UK}

\icmlcorrespondingauthor{Meredith Ringel Morris}{merrie@google.com}

\icmlkeywords{AI, AGI, Artificial General Intelligence, General AI, Human-Level AI, HLAI, ASI, frontier models, benchmarking, metrics, AI safety, AI risk, autonomous systems, Human-AI Interaction}

\vskip 0.3in
]



\printAffiliationsAndNotice{}  

\begin{abstract}
We propose a framework for classifying the capabilities and behavior of Artificial General Intelligence (AGI) models and their precursors. This framework introduces levels of AGI performance, generality, and autonomy, providing a common language to compare models, assess risks, and measure progress along the path to AGI. To develop our framework, we analyze existing definitions of AGI, and distill six principles that a useful ontology for AGI should satisfy. With these principles in mind, we propose “Levels of AGI” based on depth (performance) and breadth (generality) of capabilities, and reflect on how current systems fit into this ontology. We discuss the challenging requirements for future benchmarks that quantify the behavior and capabilities of AGI models against these levels. Finally, we discuss how these levels of AGI interact with deployment considerations such as autonomy and risk, and emphasize the importance of carefully selecting Human-AI Interaction paradigms for responsible and safe deployment of highly capable AI systems. 
\end{abstract}

\section{Introduction}
\label{intro}

Artificial General Intelligence (AGI) is an important and sometimes controversial concept in computing research, used to describe an AI system that is at least as capable as a human at most tasks. Given the rapid advancement of Machine Learning (ML) models, the concept of AGI has 
grown from a subject of philosophical debate, to one which also has near-term practical relevance. 
Some experts believe that “sparks” of AGI \citep{bubeck2023sparks} are already present in the latest generation of large language models (LLMs); some predict AI will broadly outperform humans within about a decade \citep{riskPaper}; some even assert that current LLMs \textit{are} AGIs \citep{blaiseAGI}. 

The concept of AGI
is important as it maps onto goals for, predictions about, and risks of AI:

\textbf{Goals}: Achieving human-level “intelligence” is an implicit or explicit north-star goal for many in our field, from the 1956 Dartmouth AI Conference \citep{dartmouthAI} that kick-started the modern field of AI, 
to today's leading AI research firms, whose mission statements include goals such as
“ensure transformative AI helps people and society” \citep{anthMission} and “ensure that artificial general intelligence benefits all of humanity” \citep{openAIMission}. 

\textbf{Predictions}: The concept of AGI is related to a prediction about progress in AI, namely that it is toward greater generality, approaching and exceeding human generality. Additionally, AGI is typically intertwined with a notion of “emergent” properties \citep{wei2022emergent}, i.e. capabilities not explicitly anticipated by the developer. Such capabilities offer promise, perhaps including abilities that are complementary to typical human skills, enabling new types of interaction or novel industries. Such predictions about AGI's capabilities in turn predict likely societal impacts; AGI may have significant economic implications, i.e., reaching the necessary criteria for widespread labor substitution \citep{laborMcK, centaurs, eloundou2023gpts}, as well as geo-political implications relating not only to the economic advantages AGI may confer, but also to military considerations \citep{kissinger}.  

\textbf{Risks}: Lastly, AGI is viewed by some as a concept for identifying the point when there are extreme risks \citep{shevlane2023model, riskPaper}, as  some speculate that AGI systems might be able to deceive and manipulate, accumulate resources, advance goals, behave agentically, outwit humans in broad domains, displace humans from key roles, and/or recursively self-improve.

\textbf{In this position paper, we argue that it is critical for the AI research community to explicitly reflect on what we mean by  ``AGI," and aspire to quantify attributes like the performance, generality, and autonomy of AI systems.} 
Shared operationalizable definitions for these concepts will support:
comparisons between models; risk assessments and mitigation strategies; clear criteria from policymakers and regulators; identifying goals, predictions, and risks for research and development; and the ability to understand and communicate where we are along the path to AGI.

\section{Defining AGI: Case Studies}
\label{cases}

Many AI researchers and organizations have proposed definitions of AGI. In this section, we consider nine prominent examples, and reflect on their strengths and limitations. This analysis informs our subsequent introduction of a two-dimensional, leveled ontology of AGI. 

\textbf{Case Study 1: The Turing Test.} The Turing Test \citep{turingTest} is perhaps the most well-known attempt to operationalize an AGI-like concept. Turing’s “imitation game” 
attempts 
to operationalize the question of whether machines can think, and asks a human to interactively distinguish whether text is produced by another human or by a machine. The test as originally framed is a thought experiment, and is the subject of many critiques \citep{turingWeaknesses}; in practice, the test often highlights the ease of fooling people \citep{eliza, goostman} rather than the “intelligence” of the machine. Given that modern LLMs pass some framings of the Turing Test, it seems clear that this criteria is insufficient for operationalizing or benchmarking AGI. We agree with Turing that whether a machine can think, while an interesting philosophical and scientific question, seems orthogonal to the question of what the machine can do; the latter is much more straightforward to measure and more important for evaluating impacts. Therefore we propose that AGI should be defined in terms of \textit{capabilities} rather than \textit{processes}\footnote{As research into mechanistic interpretability \citep{transparent} advances, it may enable process-oriented metrics. These may be relevant to future definitions of AGI.}.

\textbf{Case Study 2: Strong AI -- Systems Possessing Consciousness}. Philosopher John Searle mused, ``according to strong AI, the computer is not merely a tool in the study of the mind; rather, the appropriately programmed computer really is a mind, in the sense that computers given the right programs can be literally said to understand and have other cognitive states" \citep{searle_1980}. While strong AI might be one path to achieving AGI, there is no scientific consensus on methods for determining whether machines possess strong AI attributes such as consciousness \citep{butlin2023consciousness}, making this process-oriented framing impractical.  

\textbf{Case Study 3: Analogies to the Human Brain.} The original use of the term ``artificial general intelligence" was in a 1997 article about military technologies by Mark Gubrud \citep{gubrudAGI}, which defined AGI as “AI systems that rival or surpass the human brain in complexity and speed, that can acquire, manipulate and reason with general knowledge, and that are usable in essentially any phase of industrial or military operations where a human intelligence would otherwise be needed.” This early definition emphasizes processes (rivaling the human brain in complexity) in addition to capabilities; while neural network architectures underlying modern ML systems are loosely inspired by the human brain, the success of transformer-based architectures \citep{vaswani2023attention} whose performance is not reliant on human-like learning suggests that strict brain-based processes and benchmarks are not inherently necessary for AGI. 

\textbf{Case Study 4: Human-Level Performance on Cognitive Tasks.} Legg \citep{leggThesis} and Goertzel \citep{goertzel} popularized the term AGI among computer scientists in 2001 \citep{shaneTweet}, describing AGI as a machine that is able to do the cognitive tasks that people can typically do. This definition notably focuses on non-physical tasks (i.e., not requiring robotic embodiment as a precursor to AGI). Like many definitions of AGI, this framing presents ambiguity around choices such as “what tasks?” and “which people?”.

\textbf{Case Study 5: Ability to Learn Tasks.} In \textit{The Technological Singularity} \citep{murrayBook}, Shanahan suggests that AGI is “artificial intelligence that is not specialized to carry out specific tasks, but can learn to perform as broad a range of tasks as a human.” An important property of this framing is its 
inclusion of metacognitive capabilities (learning) as a requirement for AGI.

\textbf{Case Study 6: Economically Valuable Work.} OpenAI's charter defines AGI as “highly autonomous systems that outperform humans at most economically valuable work” \citep{openAICharter}. This definition has strengths per the “capabilities, not processes” criteria, as it focuses on performance agnostic to underlying mechanisms; further, this definition offers a potential yardstick for measurement, i.e., economic value. A shortcoming of this definition is that it does not capture all of the criteria that may be part of “general intelligence.” There are tasks associated with intelligence that may not have a well-defined economic value (e.g., artistic creativity or emotional intelligence). Such properties may be indirectly accounted for in economic measures (e.g., artistic creativity might produce books or movies, emotional intelligence might relate to the ability to be a successful CEO), though whether economic value captures the full spectrum of “intelligence” remains unclear. Another challenge with framing AGI in terms of economic value is the implied need for \textit{deployment} in order to realize that value, whereas a focus on capabilities might only require the \textit{potential} for an AGI to execute a task. We may develop systems that are technically capable of performing economically important tasks but don't realize that economic value for varied reasons (legal, ethical, social, etc.).

\textbf{Case Study 7: Flexible and General -- The ``Coffee Test" and Related Challenges.} Marcus suggests that AGI is “shorthand for any intelligence (there might be many) that is flexible and general, with resourcefulness and reliability comparable to (or beyond) human intelligence” \citep{marcusTwitter}. This definition captures both \textit{generality} and \textit{performance} (via the inclusion of reliability); the mention of “flexibility” is noteworthy, since, like the Shanahan formulation, this suggests that \textit{metacognitive} capabilities, such as the ability to learn new skills, are necessary to make an AI system sufficiently general. 
Further, Marcus proposes five tasks to gauge success (understanding a movie, understanding a novel, cooking in an arbitrary kitchen, writing a bug-free 10,000 line program, and converting natural language mathematical proofs into symbolic form) \citep{marcusBlog}. Accompanying a definition with a benchmark is valuable; however, more work would be required to 
make this benchmark comprehensive.
While failing some of these tasks may indicate a system is \textit{not} an AGI, it is unclear that passing them is sufficient for AGI status. In \cref{testing}, we further discuss the challenge in developing a set of tasks that is both necessary and sufficient for capturing the generality of AGI. We also note that one of Marcus’ proposed tasks, “work as a competent cook in an arbitrary kitchen” (a variant of Steve Wozniak’s “Coffee Test” \citep{wozVideo}), requires robotic embodiment; this differs from other definitions that focus on non-physical tasks\footnote{Though robotics might also be implied by the OpenAI charter's focus on “economically valuable work,” OpenAI shut down its robotics research division in 2021 \citep{roboticsCut}, suggesting this is not their intended interpretation.}. 

\textbf{Case Study 8: Artificial Capable Intelligence.} Suleyman proposed the concept of ``Artificial Capable Intelligence (ACI)" \citep{comingWave} to refer to AI systems with sufficient performance and generality to accomplish complex, multi-step tasks in the open world. More specifically, Suleyman proposed an economically-based definition of ACI skill that he dubbed the “Modern Turing Test,” in which an AI would be given \$100,000 of capital and tasked with turning that into \$1,000,000 over a period of several months. This framing is more narrow than OpenAI’s definition of economically valuable work and has the additional downside of potentially introducing alignment risks \citep{kenton2021alignment} by only targeting fiscal profit. However, a strength of Suleyman’s concept is the focus on performing a complex, multi-step task that humans value. Construed more broadly than making a million dollars, ACI's emphasis on complex, real-world tasks is noteworthy, since such tasks may have more \emph{ecological validity} than many current AI benchmarks; Marcus' aforementioned five tests of flexibility and generality \citep{marcusBlog} seem within the spirit of ACI, as well.

\textbf{Case Study 9: SOTA LLMs as Generalists.} Ag\"uera y Arcas and Norvig \citep{blaiseAGI} suggested that state-of-the-art LLMs (e.g. mid-2023 deployments of GPT-4, Bard, Llama 2, and Claude) already \textit{are} AGIs, arguing that \textit{generality} is the key property of AGI, and that because language models can discuss a wide range of topics, execute a wide range of tasks, handle multimodal inputs and outputs, operate in multiple languages, and “learn” from zero-shot or few-shot examples, they have achieved sufficient generality. While we agree that generality is a crucial characteristic of AGI, we posit that it must also be paired with a measure of \textit{performance} (i.e., if an LLM can write code or perform math, but is not reliably correct, then its generality is not yet sufficiently performant).

\section{Defining AGI: Six Principles}
\label{principles}

Reflecting on these nine example formulations of AGI (or AGI-adjacent concepts), we identify properties and commonalities that we feel contribute to a clear, operationalizable definition of AGI. We argue that any definition of AGI should meet the following six criteria:

\textbf{1. Focus on Capabilities, not Processes.} The majority of definitions focus on what an AGI can accomplish, not on the mechanism by which it accomplishes tasks. This is important for identifying characteristics that are not necessarily a prerequisite for achieving AGI (but may nonetheless be interesting research topics). This focus on capabilities implies that AGI systems need not necessarily \textit{think} or \textit{understand} in a human-like way (since this focuses on processes); similarly, it is not a necessary precursor for AGI that systems possess qualities such as \textit{consciousness} (subjective awareness) \citep{butlin2023consciousness} or \textit{sentience} (the ability to have feelings), since these qualities have a process focus.

\textbf{2. Focus on Generality \textit{and} Performance.} All of the above definitions emphasize generality to varying degrees, but some exclude performance criteria. We argue that both generality and performance are key components of AGI. In \cref{levels} we introduce a leveled taxonomy that considers the interplay between these dimensions.

\textbf{3. Focus on Cognitive and Metacognitive, but not Physical, Tasks.} Whether to require robotic embodiment \citep{roy2021machine} as a criterion for AGI is a matter of some debate. Most definitions focus on cognitive tasks, by which we mean non-physical tasks. Despite recent advances in robotics \citep{brohan2023rt2}, physical capabilities for AI systems seem to be lagging behind non-physical capabilities. It is possible that embodiment in the physical world is necessary for building the world knowledge to be successful on some cognitive tasks \citep{murrayEmbody}, or at least may be one path to success on some classes of cognitive tasks; if that turns out to be true then embodiment may be critical to some paths toward AGI. We suggest that the ability to perform physical tasks increases a system’s generality, but should not be considered a necessary prerequisite to achieving AGI. On the other hand, metacognitive capabilities (such as the ability to learn new tasks or the ability to know when to ask for clarification or assistance from a human) are key prerequisites for systems to achieve generality.  

\textbf{4. Focus on Potential, not Deployment.} Demonstrating that a system can perform a requisite set of tasks at a given level of performance should be sufficient for declaring the system to be an AGI; deployment of such a system in the open world should not be inherent in the definition of AGI. For instance, defining AGI in terms of reaching a certain level of labor substitution would require real-world deployment, whereas defining AGI in terms of being \textit{capable} of substituting for labor would focus on potential. Requiring deployment as a condition of measuring AGI introduces non-technical hurdles such as legal and social considerations, as well as ethical and safety concerns. 

\textbf{5. Focus on Ecological Validity.} Tasks that can be used to benchmark progress toward AGI are critical to operationalizing any proposed definition. While we discuss this further in \cref{testing}, we emphasize here the importance of choosing tasks that align with real-world (i.e., ecologically valid) tasks that people value (construing “value” broadly, not only as economic value but also social value, artistic value, etc.). This may mean eschewing traditional AI metrics that are easy to automate or quantify \citep{raji2021ai} but may not capture the skills that people would value in an AGI. 

\textbf{6. Focus on the Path to AGI, not a Single Endpoint.} Much as the adoption of a standard set of Levels of Driving Automation \citep{drivingLevels} allowed for clear discussions of policy and progress relating to autonomous vehicles, we posit there is value in defining “Levels of AGI.”  As we discuss in \cref{testing} and \cref{risk}, we intend for each level of AGI to be associated with a clear set of metrics/benchmarks, as well as identified risks introduced at each level, and resultant changes to the Human-AI Interaction paradigm \citep{morris2023design}. This level-based approach to defining AGI supports the coexistence of many prominent formulations – for example, Aguera y Arcas \& Norvig’s definition \citep{blaiseAGI} would fall into the “Emerging AGI” category of our ontology, while OpenAI’s threshold of labor replacement \citep{openAICharter} better matches “Exceptional AGI.” Our “Competent AGI” level is probably the best catch-all for many existing definitions of AGI (e.g., the Legg \citep{leggThesis}, Shanahan \citep{murrayBook}, and Suleyman \citep{comingWave} formulations). In the next section, we introduce a level-based ontology of AGI.    

\section{Levels of AGI}
\label{levels}

\begin{table*}[t]
\caption{A leveled, matrixed approach toward classifying systems on the path to AGI based on depth (performance) and breadth (generality) of capabilities. The assignment of example systems to cells is approximate.
Unambiguous classification of AI systems will require a standardized benchmark of tasks, as we discuss in \cref{testing}.  Note that general systems that broadly perform at a level \textit{N} may be able to perform a narrow subset of tasks at higher levels. The ``Competent AGI" level, which has not been achieved by any public systems at the time of writing, best corresponds to many prior conceptions of AGI, and may precipitate rapid societal change once achieved.}
\label{tab:table1}
\begin{center}
\begin{small}
\begin{tabular}{|p{0.31\textwidth}|p{0.31\textwidth}|p{0.31\textwidth}|}
    \hline
    \textbf{Performance} (rows) x \newline \textbf{Generality} (columns) & \textbf{Narrow} \newline \textit{clearly scoped task or set of tasks} & \textbf{General} \newline \textit{wide range of non-physical tasks, including metacognitive tasks like learning new skills} \\ 
     \specialrule{1.3pt}{0pt}{0pt}
    \textbf{Level 0: No AI} & \textbf{Narrow Non-AI} \newline calculator software; compiler & \textbf{General Non-AI} \newline human-in-the-loop computing, e.g., Amazon Mechanical Turk  
    \\ \hline
    \textbf{Level 1: Emerging} \newline \textit{equal to or somewhat better than an unskilled human} & \textbf{Emerging Narrow AI} \newline GOFAI \citep{Boden_2014}; simple rule-based systems, e.g., SHRDLU \citep{shrdlu} & \textbf{Emerging AGI} \newline ChatGPT \citep{openai2023gpt4}, Bard \citep{anil2023palm}, Llama 2 \citep{touvron2023llama}, Gemini \citep{geminiBlog} \\ \hline
    \textbf{Level 2: Competent} \newline \textit{at least 50th percentile of skilled adults} & \textbf{Competent Narrow AI} \newline  toxicity detectors such as Jigsaw \citep{das2022toxic}; Smart Speakers such as Siri \citep{siri}, Alexa \citep{alexa}, or Google Assistant \citep{gasst}; VQA systems such as PaLI \citep{chen2023pali}; Watson \citep{watson}; SOTA LLMs for a subset of tasks (e.g., short essay writing, simple coding) & \textbf{Competent AGI} \newline not yet achieved \\ \hline
    \textbf{Level 3: Expert} \newline \textit{at least 90th percentile of skilled adults} & \textbf{Expert Narrow AI} \newline spelling \& grammar checkers such as Grammarly \citep{grammarly}; generative image models such as Imagen \citep{saharia2022photorealistic} or Dall-E 2 \citep{dalle2} & \textbf{Expert AGI} \newline not yet achieved  \\ \hline
    \textbf{Level 4: Exceptional} \newline \textit{at least 99th percentile of skilled adults} & \textbf{Exceptional Narrow AI} \newline Deep Blue \citep{deepblue}, AlphaGo \citep{alphago, alphagoRL} & \textbf{Exceptional AGI} \newline not yet achieved  \\ \hline
    \textbf{Level 5: Superhuman} \newline \textit{outperforms 100\% of humans} & \textbf{Superhuman Narrow AI} \newline AlphaFold \citep{alphafold1,alphafold2}, AlphaZero \citep{alphazero}, StockFish \citep{stockfish} 
    & \textbf{Artificial Superintelligence (ASI)} \newline not yet achieved \\ \hline
    \end{tabular}
\end{small}
\end{center}
\end{table*}

In accordance with Principle 2 (``Focus on Generality and Performance") and Principle 6 (``Focus on the Path to AGI, not a Single Endpoint"), in Table \ref{tab:table1} we introduce a matrixed leveling system that focuses on \textit{performance} and \textit{generality} as the two dimensions that are core to AGI:

\textbf{Performance} refers to the \textit{depth} of an AI system’s capabilities, i.e., how it compares to human-level performance for a given task. Note that for all performance levels above “Emerging,” percentiles are in reference to a sample of adults who possess the relevant skill (e.g., “Competent” or higher performance on a task such as English writing ability would only be measured against the set of adults who are literate and fluent in English). 

\textbf{Generality} refers to the \textit{breadth} of an AI system’s capabilities, i.e., the range of tasks for which an AI system reaches a target performance threshold.

This taxonomy specifies the minimum performance over most tasks needed to achieve a given rating – e.g., a Competent AGI must have performance at least at the 50th percentile for skilled adult humans on most cognitive tasks, but may have Expert, Exceptional\footnote{While Level 4 was originally called “Virtuoso AGI,” we now use the term “Exceptional AGI,” \cite{shah2025approachtechnicalagisafety} which we believe better captures this capability level.}, or even Superhuman performance on a subset of tasks. As an example of how individual systems may straddle different points in our taxonomy, we posit that as of this writing in September 2023, frontier language models (e.g., ChatGPT \citep{openai2023gpt4}, Bard \citep{anil2023palm}, Llama2 \citep{touvron2023llama}, etc.) exhibit “Competent” performance levels for some tasks (e.g., short essay writing, simple coding), but are still at “Emerging” performance levels for most tasks (e.g., mathematical abilities, tasks involving factuality). Overall, current frontier language models would therefore be considered a Level 1 General AI (“Emerging AGI”) until the performance level increases for a broader set of tasks (at which point the Level 2 General AI, “Competent AGI,” criteria would be met). We suggest that documentation for frontier AI models, such as model cards \citep{Mitchell_2019}, should detail this mixture of performance levels. This will help end-users, policymakers, and other stakeholders come to a shared, nuanced understanding of the likely uneven performance of systems progressing along the path to AGI.

The order in which stronger skills in specific cognitive areas are acquired may have serious implications for AI safety (e.g., acquiring strong knowledge of chemical engineering before acquiring strong ethical reasoning skills may be a dangerous combination). Note also that the rate of progression between levels of performance and/or generality may be nonlinear. Acquiring the capability to learn new skills may particularly accelerate progress toward the next level.

While this taxonomy rates systems according to their performance, systems that are \textit{capable} of achieving a certain level of performance (e.g., against a given benchmark) may not match this level \textit{in practice} when deployed. For instance, user interface limitations may reduce deployed performance. Consider DALLE-2 \citep{dalle2}, which we estimate as a Level 3 Narrow AI (“Expert Narrow AI”) in our taxonomy. We estimate the “Expert” level of performance since DALLE-2 produces images of higher quality than most people are able to draw; however, the system has failure modes (e.g., drawing hands with incorrect numbers of digits, rendering nonsensical or illegible text) that prevent it from achieving an “Exceptional” performance designation.  While theoretically an “Expert” level system, \textit{in practice} the system may only be “Competent,” because prompting interfaces are too complex for most end-users to elicit optimal performance (as evidenced by user studies \citep{cantprompt} and the existence of marketplaces (e.g., \citet{promptbase}) in which skilled prompt engineers sell prompts). This observation emphasizes the importance of designing ecologically valid benchmarks (that approximate deployed rather than idealized performance), as well as the importance of considering the human-AI interaction paradigms. 

The highest level in our matrix in terms of combined performance and generality is ASI (Artificial Superintelligence). We define ``Superhuman" performance as outperforming 100\% of humans. For instance, we posit that AlphaFold \citep{alphafold1, alphafold2} is a Level 5 Narrow AI (``Superhuman Narrow AI") since it performs a single task (predicting a protein’s 3D structure from an amino acid sequence) above the level of the world’s top scientists. This definition means that Level 5 General AI (``ASI") systems will be able to do a wide range of tasks at a level that no human can match. Additionally, this framing also implies that Superhuman systems may be able to perform an even broader generality of tasks than lower levels of AGI, since the ability to execute tasks that qualitatively differ from existing human skills would by definition outperform all humans (who fundamentally cannot do such tasks). For example, non-human skills that an ASI might have could include capabilities such as neural interfaces (perhaps through mechanisms such as analyzing brain signals to decode thoughts \citep{neuralUI, musicBCI}), oracular abilities (perhaps through mechanisms such as analyzing large volumes of data to make high-quality predictions \citep{schoenegger2023large}), or the ability to communicate with animals (perhaps by mechanisms such as analyzing patterns in their vocalizations, brain waves, or body language \citep{goldwasser2023theory, ANDREAS2022104393}).

\section{Testing for AGI}
\label{testing}

Two of our six proposed principles for defining AGI (Principle 2: Generality \textit{and} Performance; Principle 6: Focus on the Path to AGI) influenced our choice of a matrixed, leveled ontology for facilitating nuanced discussions of the breadth and depth of AI capabilities. Our remaining four principles (Principle 1: Capabilities, not Processes; Principle 3: Cognitive and Metacognitive Tasks; Principle 4: Potential, not Deployment; and Principle 5: Ecological Validity) relate to the issue of measurement.  

While our \textit{performance} dimension specifies one aspect of measurement (e.g., percentile ranges for task performance relative to particular subsets of people), our \textit{generality} dimension leaves open important questions: What is the set of tasks that constitute the generality criteria? What proportion of such tasks must an AI system master to achieve a given level of generality in our schema? Are there some tasks that must always be performed to meet the criteria for certain generality levels, such as metacognitive tasks?

Operationalizing an AGI definition requires answering these questions, as well as developing specific diverse and challenging tasks. Because of the immense complexity of this process, as well as the importance of including a wide range of perspectives (including cross-organizational and multi-disciplinary viewpoints), we do not propose a benchmark in this paper. Instead, we work to clarify the ontology a benchmark should attempt to measure. We also discuss properties an AGI benchmark should possess. 

Our intent is that an AGI benchmark would include a broad suite of cognitive and metacognitive tasks (per Principle 3), measuring diverse properties including (but not limited to) linguistic intelligence, mathematical and logical reasoning \citep{logic}, spatial reasoning, interpersonal and intra-personal social intelligences, the ability to learn new skills \citep{chollet2019measure}, and creativity. A benchmark \textit{might} include tests covering psychometric categories proposed by theories of intelligence from psychology, neuroscience, cognitive science, and education; however, such tests must first be evaluated for suitability for benchmarking computing systems, since many may lack ecological and construct validity in this context \citep{serapio2023personality}. 

We emphasize the importance of metacognition, and suggest that an AGI benchmark should include metacognitive tasks such as (1) the ability to learn new skills, (2) the ability to know when to ask for help, and (3) social metacognitive abilities such as those relating to theory of mind. The ability to learn new skills \citep{chollet2019measure} is essential to generality, since it is infeasible for a system to be optimized for all possible use cases a priori; this necessitates related sub-skills such as the ability to select appropriate strategies for learning \citep{pressleyMetacog}. Knowing when to ask for help is necessary to support alignment and appropriate human-AI interaction \citep{terryHCIAlignment}, and would include an awareness of the limits of the model’s own abilities \citep{demetriouMetacog}, which relates to the sub-skill of model calibration \citep{liang2023holistic}, i.e., the model’s ability to proactively anticipate and retroactively evaluate how well it would do/did on certain tasks. Additionally, theory of mind tasks are sometimes considered metacognitive \citep{TullisMetacog}, though are sometimes classified separately as social cognition \citep{gardnerTheory}; the ability of systems to accurately model end-users is a necessary component of alignment for AGI systems. 

One open question for benchmark design is whether to allow the use of tools, including potentially AI-powered tools, as an aid to human performance. This choice may ultimately be task dependent and should account for ecological validity in benchmark choice (per Principle 5). For example, in determining whether a self-driving car is sufficiently safe, benchmarking against a person driving without the benefit of any modern AI-assisted safety tools would not be the most informative comparison; since the relevant counterfactual involves some driver-assistance technology, we may prefer a comparison to that baseline. 

While an AGI benchmark might draw from some existing AI benchmarks \citep{haiblog} (e.g., HELM \citep{liang2023holistic}, BIG-bench \citep{srivastava2023imitation}), we also envision the inclusion of open-ended and/or interactive tasks that might require qualitative evaluation \citep{papakyriakopoulos2021qualitative, yang2023dawn, bubeck2023sparks}. We suspect that these latter classes of complex, open-ended tasks, though difficult to benchmark, will have better ecological validity than traditional AI metrics, or than adapted traditional measures of human intelligence. 

It is impossible to enumerate the full set of tasks achievable by a sufficiently general intelligence. As such, an AGI benchmark should be a \textit{living} benchmark. Such a benchmark should therefore include a framework for generating and agreeing upon new tasks. 

Determining that something is \textit{not} an AGI at a given level simply requires identifying tasks that people can typically do but the system cannot adequately perform. Systems that pass the majority of the envisioned AGI benchmark at a particular performance level (``Emerging," ``Competent," etc.), including new tasks added by the testers, can be assumed to have the associated level of generality for practical purposes (i.e., though in theory there could still be a test the AGI would fail, at some point unprobed failures are so specialized or atypical as to be practically irrelevant). 
We hesitate to specify the number or percentage of tasks that a system must pass at a given level of performance in order to be declared a General AI at that Level (e.g., a rule such as ``a system must pass at least 90\% of an AGI benchmark at a given performance level to get that rating"). While we think this will be a very high percentage, it will probably not be 100\%, since it seems clear that broad but imperfect generality is impactful (individual humans also lack consistent performance across all possible tasks, but are generally intelligent). Determining what portion of benchmarking tasks at a given level demonstrate generality remains an open research question.  


\section{Risk, Autonomy, and Interaction}
\label{risk}

Discussions of AGI often include discussion of risk, including ``x-risk" -- existential \citep{xriskPetition} or other very extreme risks \citep{shevlane2023model}. A leveled approach to defining AGI enables a more nuanced discussion of how different combinations of performance and generality relate to different types of AI risk. While there is value in considering extreme risk scenarios, understanding AGI via our proposed ontology rather than as a single endpoint (per Principle 6) can help ensure that policymakers also identify and prioritize risks in the near-term and on the path to AGI. 

\subsection{Levels of AGI as a Framework for Risk Assessment}
\label{riskFramework}

As we advance along our capability levels toward ASI, new risks are introduced, including misuse risks, alignment risks, and structural risks \citep{zwetsloot2019thinking}. For example, the “Expert AGI” level is likely to involve structural risks related to economic disruption and job displacement, as more and more industries reach the substitution threshold for machine intelligence in lieu of human labor. On the other hand, reaching “Expert AGI” likely alleviates some risks introduced by “Emerging AGI” and “Competent AGI,” such as the risk of incorrect task execution. The “Exceptional AGI” and “ASI” levels are where many concerns relating to x-risk are most likely to emerge (e.g., an AI that can outperform its human operators on a broad range of tasks might deceive them to achieve a mis-specified goal, as in misalignment thought experiments \citep{alignment}).

Systemic risks such as destabilization of international relations may be a concern if the rate of progression between levels outpaces regulation or diplomacy (e.g., the first nation to achieve ASI may have a substantial geopolitical/military advantage, creating complex structural risks). At levels below “Expert AGI” (e.g., “Emerging AGI,” “Competent AGI,” and all “Narrow” AI categories), risks likely stem more from human actions (e.g., risks of AI misuse, whether accidental, incidental, or malicious). A more complete analysis of risk profiles associated with each level is a critical step toward developing a taxonomy of AGI that can guide safety/ethics research and policymaking. 

Whether an AGI benchmark should include tests for potentially dangerous capabilities (e.g., the ability to deceive, to persuade \citep{subliminal}, or to perform advanced biochemistry \citep{morris2023scientists}) is controversial. We lean on the side of including such capabilities in benchmarking, since most such skills tend to be dual use (having valid applications to socially positive scenarios as well as nefarious ones). Dangerous capability benchmarking can be de-risked via Principle 4 (Potential, not Deployment) by ensuring benchmarks for any dangerous or dual-use tasks are appropriately sandboxed and not defined in terms of deployment. However, including such tests in a public benchmark may allow malicious actors to optimize for these abilities; understanding how to mitigate risks associated with benchmarking dual-use abilities remains an important area for research by AI safety, AI ethics, and AI governance experts.

Concurrent with this work, Anthropic released Version 1.0 of its Responsible Scaling Policy (RSP) \citep{rsp}. This policy uses a levels-based approach (inspired by biosafety levels \citep{richmond2009biosafety}) to define the level of risk associated with an AI system, identifying what dangerous capabilities may be associated with each AI Safety Level (ASL), and what containment or deployment measures should be taken at each level. Current SOTA generative AIs are classified as an ASL-2 risk. Including items matched to ASL capabilities in any AGI benchmark would connect points in our AGI taxonomy to specific risks and mitigations.

\subsection{Capabilities vs. Autonomy}
\label{autonomy}

\label{autonomyTable}
\begin{table*}[h!]
\caption{More capable AI systems unlock new human-AI interaction paradigms. The choice of appropriate autonomy level need not be the maximum achievable given the capabilities of the underlying model. One consideration in the choice of autonomy level are resulting risks. This table's examples illustrate the importance of carefully considering human-AI interaction design decisions.}
\label{tab:table2}
\begin{center}
\begin{small}
\begin{tabular}{|p{0.20\textwidth}|p{0.3\textwidth}|p{0.2\textwidth}|p{.2\textwidth}|}
    \hline
    \textbf{Autonomy Level} & \textbf{Example Systems} & \textbf{Unlocking \newline AGI Level(s)} & \textbf{Example Risks \newline Introduced} \\ \specialrule{1.3pt}{0pt}{0pt}
    \textbf{Autonomy Level 0: \newline No AI} \newline \textit{human does everything} & Analogue approaches (e.g., sketching with pencil on paper) \newline\newline Non-AI digital workflows (e.g., typing in a text editor; drawing in a paint program) & No AI & n/a (status quo risks) \\ \hline
    \textbf{Autonomy Level 1: \newline AI as a Tool} \newline \textit{human fully controls task and uses AI to automate mundane sub-tasks} & Information-seeking with the aid of a search engine  \newline\newline Revising writing with the aid of a grammar-checking program \newline\newline Reading a sign with a \newline machine translation app & Possible: \newline Emerging Narrow AI \newline\newline Likely: \newline Competent Narrow AI & de-skilling \newline (e.g., over-reliance) \newline\newline disruption of \newline established \newline industries \\ \hline
    \textbf{Autonomy Level 2: \newline AI as a Consultant} \newline \textit{AI takes on a \newline substantive role, but only when invoked by a human} & Relying on a language model to summarize a set of documents \newline\newline Accelerating computer programming with a code-generating model \newline\newline Consuming most entertainment via a sophisticated recommender system & Possible: \newline Competent Narrow AI \newline\newline Likely: \newline Expert Narrow AI; \newline Emerging AGI & 
        over-trust \newline\newline 
        radicalization \newline\newline
        targeted \newline manipulation
        \\ \hline
    \textbf{Autonomy Level 3: \newline AI as a \newline Collaborator} \newline \textit{co-equal human-AI collaboration; interactive coordination of goals \& tasks} & Training as a chess player through interactions with and analysis of a chess-playing AI \newline\newline Entertainment via social interactions with AI-generated personalities & Possible: \newline Emerging AGI \newline\newline Likely: \newline Expert Narrow AI; \newline Competent AGI & anthropomorphization (e.g., parasocial \newline relationships)\newline\newline rapid societal change \\ \hline
    \textbf{Autonomy Level 4: \newline AI as an Expert} \newline \textit{AI drives interaction; human provides guidance \& feedback or performs subtasks} & Using an AI system to advance scientific discovery (e.g., protein-folding) & Possible: \newline Exceptional Narrow AI  \newline\newline Likely: \newline Expert AGI & societal-scale ennui \newline\newline mass labor \newline displacement \newline\newline decline of human exceptionalism \\ \hline
    \textbf{Autonomy Level 5: \newline AI as an Agent} \newline \textit{fully autonomous AI} & Autonomous AI-powered \newline personal assistants \newline \textit{(not yet unlocked)} & Likely: \newline Exceptional AGI; \newline ASI & misalignment \newline\newline concentration \newline of power \\ \hline
    \end{tabular}
\end{small}
\end{center}
\end{table*}

While capabilities provide prerequisites for AI risks, AI systems (including AGI systems) do not and will not operate in a vacuum. Rather, AI systems are deployed with particular interfaces and used to achieve particular tasks in specific scenarios. These contextual attributes (interface, task, scenario, end-user) have substantial bearing on risk. 

Consider, for instance, the affordances of user interfaces for AGI systems. Increasing capabilities unlock new interaction paradigms, but \textit{do not determine them}. Rather, system designers and end-users will settle on a mode of human-AI interaction \citep{morris2023design} that balances a variety of considerations, including safety. We propose characterizing human-AI interaction paradigms with six \textbf{Levels of Autonomy}, described in Table \ref{tab:table2}. 

These Levels of Autonomy are correlated with the Levels of AGI. Higher levels of autonomy are “unlocked” by AGI capability progression, though lower levels of autonomy may be desirable for particular tasks and contexts even as we reach higher levels of AGI. Carefully considered choices around human-AI interaction are vital to safe and responsible deployment of frontier AI models. 

Unlike prior taxonomies of computer automation \citep{sheridan1978human, auto8levels, auto10levels} that take a computer-centric perspective (framing automation in terms of how much control the designer relinquishes to computers), we characterize the concept of autonomy through the lens of the nature of human-AI interaction style; further, our ontology considers how AI capabilities may enable particular interaction paradigms and how the combination of level of autonomy and level of AGI may impact risk. Shneiderman \citep{shneiderman2020humancentered} observes that automation is not a zero-sum game, and that high levels of automation can co-exist with high levels of human control; this view is compatible with our perspective of considering automation through the perspective of varying styles of human-AI partnerships.  

We emphasize the importance of the “No AI” paradigm for many contexts, including for education, enjoyment, assessment, or safety reasons. For example, in the domain of self-driving vehicles, when Level 5 Self-Driving technology is widely available, there may be reasons for using a Level 0 (No Automation) vehicle. These include for instructing a new driver (education), for pleasure by driving enthusiasts (enjoyment), for driver’s licensing exams (assessment), or in conditions where sensors cannot be relied upon such as technology failures or extreme weather events (safety). While Level 5 Self-Driving \citep{drivingLevels} vehicles would likely be a Level 4 or 5 Narrow AI under our taxonomy,
the same considerations regarding human vs. computer autonomy apply to AGIs. We may develop an AGI, but choose not to deploy it autonomously, or choose to deploy it with differentiated autonomy levels in distinct circumstances as dictated by contextual considerations. 

Certain aspects of generality may be required to make particular interaction paradigms desirable. For example, the Autonomy Levels 3, 4, and 5 (``Collaborator," ``Expert," and ``Agent") may only work well if an AI system also demonstrates strong performance on certain metacognitive abilities (learning when to ask a human for help, theory of mind modeling, social-emotional skills). Implicit in our definition of Autonomy Level 5 (``AI as an Agent") is that such a fully autonomous AI can act in an aligned fashion without continuous human oversight, but knows when to consult humans \citep{shah2021benefits}. Interfaces that support human-AI alignment through better task specification, the bridging of process gulfs, and evaluation of outputs \citep{terryHCIAlignment} are a vital area of research.

\subsection{Human-AI Interaction and Risk Assessment}
\label{hai}

Table \ref{tab:table2} illustrates the interplay between AGI Level, Autonomy Level, and risk. Advances in model performance and generality unlock additional interaction paradigm choices (including full autonomy). These interaction paradigms in turn introduce new classes of risk. The interplay of model capabilities and interaction design will enable more nuanced risk assessments and responsible deployment decisions than considering model capabilities alone.

Table \ref{tab:table2} also provides concrete examples of each of our six proposed Levels of Autonomy. For each level of autonomy, we indicate the corresponding levels of performance and generality that ``unlock" that interaction paradigm (i.e., the level of AGI at which it is possible or likely for that paradigm to be successfully deployed and adopted). 

Our predictions regarding ``unlocking" levels tend to require higher levels of performance for Narrow than for General AI systems; for instance, we posit that the use of AI as a Consultant is likely with either an Expert Narrow AI or an Emerging AGI. This discrepancy reflects the fact that for General systems, capability development is likely to be uneven; for example, a Level 1 General AI (``Emerging AGI") may have Level 2 or perhaps even Level 3 performance across some subset of tasks. Such unevenness of capability for General AIs may unlock higher autonomy levels for particular tasks that are aligned with their specific strengths. 

Considering AGI systems in the context of use by people allows us to reflect on the interplay between advances in models and advances in human-AI interaction paradigms. The role of model building research can be seen as helping systems’ capabilities progress along the path to AGI in their performance and generality, such that an AI system’s abilities will overlap an increasingly large portion of human abilities. Conversely, the role of human-AI interaction research can be viewed as ensuring new AI systems are \textit{usable}  by and \textit{useful} to people such that AI systems successfully extend people’s capabilities (i.e., ``intelligence augmentation" \citep{brynjolfsson2022turing, ia_Englebart}).   

\section{Conclusion}

Artificial General Intelligence is a concept of both aspirational and practical consequences. We analyzed nine definitions of AGI, identifying strengths and weaknesses. Based on this analysis, we introduced six principles for a clear, operationalizable definition of AGI: focusing on capabilities, not processes; focusing on generality \textit{and} performance; focusing on cognitive and metacognitive (rather than physical) tasks; focusing on potential rather than deployment; focusing on ecological validity for benchmarking; and focusing on the path to AGI rather than a single endpoint.

With these principles in mind, we introduced our Levels of AGI ontology, which offers a more nuanced way to define progress toward AGI by considering generality (either Narrow or General) in tandem with five levels of performance (Emerging, Competent, Expert, Exceptional, and Superhuman). We reflected on how current AI systems and AGI definitions fit into this framing. Further, we discussed the implications of our principles for developing a living, ecologically valid AGI benchmark, and argue that such an endeavor, while sure to be challenging, is vital 
to engage with.  

Finally, we considered how our principles and ontology can reshape discussions around the risks associated with AGI. Notably, we observed that AGI is not necessarily synonymous with autonomy. We introduced Levels of Autonomy that are unlocked, but not determined by, progression through the Levels of AGI. We illustrated how considering AGI Level jointly with Autonomy Level can provide more nuanced insights into risks associated with AI systems, underscoring the importance of investing in human-AI interaction research in tandem with model improvements. 

We hope our framework will prove adaptable and scalable -- for instance, how we define and measure progress toward AGI might change with technical advances such as improvements in interpretability that provide insight into models' inner workings. Additionally, parts of our ontology such as Human-AI Interaction paradigms and associated risks might evolve as society itself adapts to advances in AI. 

\section*{Impact Statement}
This position paper introduces a novel ontology that supports discussing progress toward AGI in a nuanced manner, with the aim of supporting clear communication among researchers, practitioners, and policymakers about systems' capabilities and associated risks. 

\section*{Acknowledgements}

Thank you to the members of the Google DeepMind PAGI team for their support of this effort, and to Martin Wattenberg, Michael Terry, Geoffrey Irving, Murray Shanahan, Dileep George, Blaise Agüera y Arcas, and Ben Shneiderman for helpful discussions about this topic.

\bibliography{levels_of_AGI}

\begin{thebibliography}{96}
\providecommand{\natexlab}[1]{#1}
\providecommand{\url}[1]{\texttt{#1}}
\expandafter\ifx\csname urlstyle\endcsname\relax
  \providecommand{\doi}[1]{doi: #1}\else
  \providecommand{\doi}{doi: \begingroup \urlstyle{rm}\Url}\fi

\bibitem[{Ag\"uera y Arcas} \& Norvig(2023){Ag\"uera y Arcas} and Norvig]{blaiseAGI}
{Ag\"uera y Arcas}, B. and Norvig, P.
\newblock {Artificial General Intelligence is Already Here}.
\newblock Noema, October 2023.
\newblock URL \url{https://www.noemamag.com/artificial-general-intelligence-is-already-here/}.

\bibitem[Amazon()]{alexa}
Amazon.
\newblock {Amazon Alexa}.
\newblock URL \url{https://alexa.amazon.com/}.
\newblock accessed on October 20, 2023.

\bibitem[Andreas et~al.(2022)Andreas, Beguš, Bronstein, Diamant, Delaney, Gero, Goldwasser, Gruber, {de Haas}, Malkin, Pavlov, Payne, Petri, Rus, Sharma, Tchernov, Tønnesen, Torralba, Vogt, and Wood]{ANDREAS2022104393}
Andreas, J., Beguš, G., Bronstein, M.~M., Diamant, R., Delaney, D., Gero, S., Goldwasser, S., Gruber, D.~F., {de Haas}, S., Malkin, P., Pavlov, N., Payne, R., Petri, G., Rus, D., Sharma, P., Tchernov, D., Tønnesen, P., Torralba, A., Vogt, D., and Wood, R.~J.
\newblock Toward understanding the communication in sperm whales.
\newblock \emph{iScience}, 25\penalty0 (6):\penalty0 104393, 2022.
\newblock ISSN 2589-0042.
\newblock \doi{https://doi.org/10.1016/j.isci.2022.104393}.
\newblock URL \url{https://www.sciencedirect.com/science/article/pii/S2589004222006642}.

\bibitem[Anil et~al.(2023)Anil, Dai, Firat, and et~al.]{anil2023palm}
Anil, R., Dai, A.~M., Firat, O., and et~al.
\newblock {PaLM 2 Technical Report}.
\newblock \emph{CoRR}, abs/2305.10403, 2023.
\newblock \doi{10.48550/arXiv.2305.10403}.
\newblock URL \url{https://arxiv.org/abs/2305.10403}.

\bibitem[Anthropic(2023{\natexlab{a}})]{anthMission}
Anthropic.
\newblock {Company: Anthropic}, 2023{\natexlab{a}}.
\newblock URL \url{https://www.anthropic.com/company}.
\newblock Accessed October 12, 2023.

\bibitem[Anthropic(2023{\natexlab{b}})]{rsp}
Anthropic.
\newblock {Anthropic's Responsible Scaling Policy}, September 2023{\natexlab{b}}.
\newblock URL \url{https://www-files.anthropic.com/production/files/responsible-scaling-policy-1.0.pdf}.
\newblock accessed on October 20, 2023.

\bibitem[Apple()]{siri}
Apple.
\newblock Siri.
\newblock URL \url{https://www.apple.com/siri/}.
\newblock accessed on October 20, 2023.

\bibitem[Bellier et~al.(2023)Bellier, Llorens, Marciano, Gunduz, Schalk, Brunner, and Knight]{musicBCI}
Bellier, L., Llorens, A., Marciano, D., Gunduz, A., Schalk, G., Brunner, P., and Knight, R.~T.
\newblock Music can be reconstructed from human auditory cortex activity using nonlinear decoding models.
\newblock \emph{PLOS Biology}, 21\penalty0 (8):\penalty0 1--27, 08 2023.
\newblock \doi{10.1371/journal.pbio.3002176}.
\newblock URL \url{https://doi.org/10.1371/journal.pbio.3002176}.

\bibitem[Bengio et~al.(2023)Bengio, Hinton, Yao, Song, Abbeel, Harari, Zhang, Xue, Shalev-Shwartz, Hadfield, Clune, Maharaj, Hutter, Baydin, McIlraith, Gao, Acharya, Krueger, Dragan, Torr, Russell, Kahneman, Brauner, and Mindermann]{riskPaper}
Bengio, Y., Hinton, G., Yao, A., Song, D., Abbeel, P., Harari, Y.~N., Zhang, Y.-Q., Xue, L., Shalev-Shwartz, S., Hadfield, G., Clune, J., Maharaj, T., Hutter, F., Baydin, A.~G., McIlraith, S., Gao, Q., Acharya, A., Krueger, D., Dragan, A., Torr, P., Russell, S., Kahneman, D., Brauner, J., and Mindermann, S.
\newblock {Managing AI Risks in an Era of Rapid Progress}.
\newblock \emph{CoRR}, abs/2310.17688, 2023.
\newblock \doi{10.48550/arXiv.2310.17688}.
\newblock URL \url{https://arxiv.org/abs/2310.17688}.

\bibitem[Boden(2014)]{Boden_2014}
Boden, M.~A.
\newblock \emph{GOFAI}, pp.\  89–107.
\newblock Cambridge University Press, 2014.

\bibitem[Brohan et~al.(2023)Brohan, Brown, Carbajal, Chebotar, Chen, Choromanski, Ding, Driess, Dubey, Finn, Florence, Fu, Arenas, Gopalakrishnan, Han, Hausman, Herzog, Hsu, Ichter, Irpan, Joshi, Julian, Kalashnikov, Kuang, Leal, Lee, Lee, Levine, Lu, Michalewski, Mordatch, Pertsch, Rao, Reymann, Ryoo, Salazar, Sanketi, Sermanet, Singh, Singh, Soricut, Tran, Vanhoucke, Vuong, Wahid, Welker, Wohlhart, Wu, Xia, Xiao, Xu, Xu, Yu, and Zitkovich]{brohan2023rt2}
Brohan, A., Brown, N., Carbajal, J., Chebotar, Y., Chen, X., Choromanski, K., Ding, T., Driess, D., Dubey, A., Finn, C., Florence, P., Fu, C., Arenas, M.~G., Gopalakrishnan, K., Han, K., Hausman, K., Herzog, A., Hsu, J., Ichter, B., Irpan, A., Joshi, N., Julian, R., Kalashnikov, D., Kuang, Y., Leal, I., Lee, L., Lee, T.-W.~E., Levine, S., Lu, Y., Michalewski, H., Mordatch, I., Pertsch, K., Rao, K., Reymann, K., Ryoo, M., Salazar, G., Sanketi, P., Sermanet, P., Singh, J., Singh, A., Soricut, R., Tran, H., Vanhoucke, V., Vuong, Q., Wahid, A., Welker, S., Wohlhart, P., Wu, J., Xia, F., Xiao, T., Xu, P., Xu, S., Yu, T., and Zitkovich, B.
\newblock {RT-2: Vision-Language-Action Models Transfer Web Knowledge to Robotic Control}.
\newblock \emph{CoRR}, abs/2307.15818, 2023.
\newblock \doi{10.48550/arXiv.2307.15818}.
\newblock URL \url{https://arxiv.org/abs/2307.15818}.

\bibitem[Brynjolfsson(2022)]{brynjolfsson2022turing}
Brynjolfsson, E.
\newblock {The Turing Trap: The Promise \& Peril of Human-Like Artificial Intelligence}.
\newblock \emph{CoRR}, abs/2201.04200, 2022.
\newblock \doi{10.48550/arXiv.2201.04200}.
\newblock URL \url{https://arxiv.org/abs/2201.04200}.

\bibitem[Bubeck et~al.(2023)Bubeck, Chandrasekaran, Eldan, Gehrke, Horvitz, Kamar, Lee, Lee, Li, Lundberg, Nori, Palangi, Ribeiro, and Zhang]{bubeck2023sparks}
Bubeck, S., Chandrasekaran, V., Eldan, R., Gehrke, J., Horvitz, E., Kamar, E., Lee, P., Lee, Y.~T., Li, Y., Lundberg, S., Nori, H., Palangi, H., Ribeiro, M.~T., and Zhang, Y.
\newblock {Sparks of Artificial General Intelligence: Early experiments with GPT-4}.
\newblock \emph{CoRR}, abs/2303.12712, 2023.
\newblock \doi{10.48550/arXiv.2303.12712}.
\newblock URL \url{https://arxiv.org/abs/2303.12712}.

\bibitem[Butlin et~al.(2023)Butlin, Long, Elmoznino, Bengio, Birch, Constant, Deane, Fleming, Frith, Ji, Kanai, Klein, Lindsay, Michel, Mudrik, Peters, Schwitzgebel, Simon, and VanRullen]{butlin2023consciousness}
Butlin, P., Long, R., Elmoznino, E., Bengio, Y., Birch, J., Constant, A., Deane, G., Fleming, S.~M., Frith, C., Ji, X., Kanai, R., Klein, C., Lindsay, G., Michel, M., Mudrik, L., Peters, M. A.~K., Schwitzgebel, E., Simon, J., and VanRullen, R.
\newblock {Consciousness in Artificial Intelligence: Insights from the Science of Consciousness}.
\newblock \emph{CoRR}, abs/2308.08708, 2023.
\newblock \doi{10.48550/arXiv.2308.08708}.
\newblock URL \url{https://arxiv.org/abs/2308.08708}.

\bibitem[Campbell et~al.(2002)Campbell, Hoane, and Hsu]{deepblue}
Campbell, M., Hoane, A.~J., and Hsu, F.-h.
\newblock {Deep Blue}.
\newblock \emph{Artif. Intell.}, 134\penalty0 (1–2):\penalty0 57–83, jan 2002.
\newblock ISSN 0004-3702.
\newblock \doi{10.1016/S0004-3702(01)00129-1}.
\newblock URL \url{https://doi.org/10.1016/S0004-3702(01)00129-1}.

\bibitem[Chen et~al.(2023)Chen, Wang, Changpinyo, and et~al.]{chen2023pali}
Chen, X., Wang, X., Changpinyo, S., and et~al.
\newblock {PaLI: A Jointly-Scaled Multilingual Language-Image Model}.
\newblock \emph{CoRR}, abs/2209.06794, 2023.
\newblock \doi{10.48550/arXiv.2209.06794}.
\newblock URL \url{https://arxiv.org/abs/2209.06794}.

\bibitem[Chollet(2019)]{chollet2019measure}
Chollet, F.
\newblock On the measure of intelligence, 2019.

\bibitem[Christian(2020)]{alignment}
Christian, B.
\newblock \emph{{The Alignment Problem}}.
\newblock W. W. Norton \& Company, 2020.

\bibitem[Das et~al.(2022)Das, Saha, and Das]{das2022toxic}
Das, M.~M., Saha, P., and Das, M.
\newblock {Which One is More Toxic? Findings from Jigsaw Rate Severity of Toxic Comments}.
\newblock \emph{CoRR}, abs/2206.13284, 2022.
\newblock \doi{10.48550/arXiv.2206.13284}.
\newblock URL \url{https://arxiv.org/abs/2206.13284}.

\bibitem[Dell'Acqua et~al.(2023)Dell'Acqua, McFowland, Mollick, Lifshitz-Assaf, Kellogg, Rajendran, Krayer, Candelon, and Lakhani]{centaurs}
Dell'Acqua, F., McFowland, E., Mollick, E.~R., Lifshitz-Assaf, H., Kellogg, K., Rajendran, S., Krayer, L., Candelon, F., and Lakhani, K.~R.
\newblock {Navigating the Jagged Technological Frontier: Field Experimental Evidence of the Effects of AI on Knowledge Worker Productivity and Quality}.
\newblock \emph{Harvard Business School Technology \& Operations Management Unit Working Paper Number 24-013}, September 2023.

\bibitem[Demetriou \& Kazi(2006)Demetriou and Kazi]{demetriouMetacog}
Demetriou, A. and Kazi, S.
\newblock Self-awareness in g (with processing efficiency and reasoning).
\newblock \emph{Intelligence}, 34:\penalty0 297--317, 2006.
\newblock \doi{https://doi.org/10.1016/j.intell.2005.10.002}.

\bibitem[Ellingrud et~al.(2023)Ellingrud, Sanghvi, Dandona, Madgavkar, Chui, White, and Hasebe]{laborMcK}
Ellingrud, K., Sanghvi, S., Dandona, G.~S., Madgavkar, A., Chui, M., White, O., and Hasebe, P.
\newblock {Generative AI and the future of work in America}.
\newblock McKinsey Institute Global Report, July 2023.
\newblock URL \url{https://www.mckinsey.com/mgi/our-research/generative-ai-and-the-future-of-work-in-america}.

\bibitem[Eloundou et~al.(2023)Eloundou, Manning, Mishkin, and Rock]{eloundou2023gpts}
Eloundou, T., Manning, S., Mishkin, P., and Rock, D.
\newblock Gpts are gpts: An early look at the labor market impact potential of large language models, 2023.

\bibitem[Englebart(1962)]{ia_Englebart}
Englebart, D.
\newblock Augmenting human intellect: A conceptual framework.
\newblock October 1962.
\newblock URL \url{https://www.dougengelbart.org/pubs/papers/scanned/Doug_Engelbart-AugmentingHumanIntellect.pdf}.

\bibitem[for AI~Safety(2023)]{xriskPetition}
for AI~Safety, C.
\newblock {Statement on AI Risk}, 2023.
\newblock URL \url{https://www.safe.ai/statement-on-ai-risk}.

\bibitem[Gardner(2011)]{gardnerTheory}
Gardner, H.~E.
\newblock \emph{Frames of Mind: The Theory of Multiple Intelligences}.
\newblock Basic Books, 2011.

\bibitem[Goertzel(2014)]{goertzel}
Goertzel, B.
\newblock {Artificial General Intelligence: Concept, State of the Art, and Future Prospects}.
\newblock \emph{Journal of Artificial General Intelligence}, 01 2014.
\newblock \doi{10.2478/jagi-2014-0001}.

\bibitem[Goldwasser et~al.(2023)Goldwasser, Gruber, Kalai, and Paradise]{goldwasser2023theory}
Goldwasser, S., Gruber, D.~F., Kalai, A.~T., and Paradise, O.
\newblock A theory of unsupervised translation motivated by understanding animal communication, 2023.

\bibitem[Google()]{gasst}
Google.
\newblock {Google Assistant, your own personal Google}.
\newblock URL \url{https://assistant.google.com/}.
\newblock accessed on October 20, 2023.

\bibitem[Grammarly(2023)]{grammarly}
Grammarly, 2023.
\newblock URL \url{https://www.grammarly.com/}.

\bibitem[Gubrud(1997)]{gubrudAGI}
Gubrud, M.
\newblock {Nanotechnology and International Security}.
\newblock \emph{Fifth Foresight Conference on Molecular Nanotechnology}, November 1997.

\bibitem[IBM()]{watson}
IBM.
\newblock {IBM Watson}.
\newblock URL \url{https://www.ibm.com/watson}.
\newblock accessed on October 20, 2023.

\bibitem[Jumper et~al.(2021)Jumper, Evans, Pritzel, Green, Figurnov, Ronneberger, Tunyasuvunakool, Bates, Žídek, Potapenko, Bridgland, Meyer, Kohl, Ballard, Cowie, Romera-Paredes, Nikolov, Jain, Adler, Back, Petersen, Reiman, Clancy, Zielinski, Steinegger, Pacholska, Berghammer, Bodenstein, Silver, Vinyals, Senior, Kavukcuoglu, Kohli, and Hassabis]{alphafold1}
Jumper, J., Evans, R., Pritzel, A., Green, T., Figurnov, M., Ronneberger, O., Tunyasuvunakool, K., Bates, R., Žídek, A., Potapenko, A., Bridgland, A., Meyer, C., Kohl, S. A.~A., Ballard, A.~J., Cowie, A., Romera-Paredes, B., Nikolov, S., Jain, R., Adler, J., Back, T., Petersen, S., Reiman, D., Clancy, E., Zielinski, M., Steinegger, M., Pacholska, M., Berghammer, T., Bodenstein, S., Silver, D., Vinyals, O., Senior, A.~W., Kavukcuoglu, K., Kohli, P., and Hassabis, D.
\newblock {Highly Accurate Protein Structure Prediction with AlphaFold}.
\newblock \emph{Nature}, 596:\penalty0 583--589, 2021.
\newblock \doi{10.1038/s41586-021-03819-2}.

\bibitem[Kenton et~al.(2021)Kenton, Everitt, Weidinger, Gabriel, Mikulik, and Irving]{kenton2021alignment}
Kenton, Z., Everitt, T., Weidinger, L., Gabriel, I., Mikulik, V., and Irving, G.
\newblock {Alignment of Language Agents}.
\newblock \emph{CoRR}, abs/2103.14659, 2021.
\newblock \doi{10.48550/arXiv.2103.14659}.
\newblock URL \url{https://arxiv.org/abs/2103.14659}.

\bibitem[Kissinger et~al.(2022)Kissinger, Schmidt, and Huttenlocher]{kissinger}
Kissinger, H., Schmidt, E., and Huttenlocher, D.
\newblock \emph{{The Age of AI}}.
\newblock Back Bay Books, November 2022.

\bibitem[Legg(2008)]{leggThesis}
Legg, S.
\newblock {Machine Super Intelligence}.
\newblock Doctoral Dissertation submitted to the Faculty of Informatics of the University of Lugano, June 2008.

\bibitem[Legg(2022)]{shaneTweet}
Legg, S.
\newblock Twitter (now "X"), May 2022.
\newblock URL \url{https://twitter.com/ShaneLegg/status/1529483168134451201}.
\newblock Accessed on October 12, 2023.

\bibitem[Liang et~al.(2023)Liang, Bommasani, Lee, and et~al.]{liang2023holistic}
Liang, P., Bommasani, R., Lee, T., and et~al.
\newblock {Holistic Evaluation of Language Models}.
\newblock \emph{CoRR}, abs/2211.09110, 2023.
\newblock \doi{10.48550/arXiv.2211.09110}.
\newblock URL \url{https://arxiv.org/abs/2211.09110}.

\bibitem[Lynch(2023)]{haiblog}
Lynch, S.
\newblock {AI Benchmarks Hit Saturation}.
\newblock Stanford Human-Centered Artificial Intelligence Blog, April 2023.
\newblock URL \url{https://hai.stanford.edu/news/ai-benchmarks-hit-saturation}.

\bibitem[Marcus(2022{\natexlab{a}})]{marcusBlog}
Marcus, G.
\newblock {Dear Elon Musk, here are five things you might want to consider about AGI}.
\newblock "Marcus on AI" Substack, May 2022{\natexlab{a}}.
\newblock URL \url{https://garymarcus.substack.com/p/dear-elon-musk-here-are-five-things?s=r}.

\bibitem[Marcus(2022{\natexlab{b}})]{marcusTwitter}
Marcus, G.
\newblock Twitter (now "X"), May 2022{\natexlab{b}}.
\newblock URL \url{https://twitter.com/GaryMarcus/status/1529457162811936768}.
\newblock Accessed on October 12, 2023.

\bibitem[McCarthy et~al.(1955)McCarthy, Minsky, Rochester, and Shannon]{dartmouthAI}
McCarthy, J., Minsky, M., Rochester, N., and Shannon, C.
\newblock {A Proposal for The Dartmouth Summer Research Project on Artificial Intelligence}.
\newblock Dartmouth Workshop, 1955.

\bibitem[Mitchell et~al.(2019)Mitchell, Wu, Zaldivar, Barnes, Vasserman, Hutchinson, Spitzer, Raji, and Gebru]{Mitchell_2019}
Mitchell, M., Wu, S., Zaldivar, A., Barnes, P., Vasserman, L., Hutchinson, B., Spitzer, E., Raji, I.~D., and Gebru, T.
\newblock {Model Cards for Model Reporting}.
\newblock In \emph{Proceedings of the Conference on Fairness, Accountability, and Transparency}. {ACM}, jan 2019.
\newblock \doi{10.1145/3287560.3287596}.
\newblock URL \url{https://doi.org/10.1145\%2F3287560.3287596}.

\bibitem[Morris(2023)]{morris2023scientists}
Morris, M.~R.
\newblock {Scientists' Perspectives on the Potential for Generative AI in their Fields}.
\newblock \emph{CoRR}, abs/2304.01420, 2023.
\newblock \doi{10.48550/arXiv.2304.01420}.
\newblock URL \url{https://arxiv.org/abs/2304.01420}.

\bibitem[Morris et~al.(2023)Morris, Cai, Holbrook, Kulkarni, and Terry]{morris2023design}
Morris, M.~R., Cai, C.~J., Holbrook, J., Kulkarni, C., and Terry, M.
\newblock {The Design Space of Generative Models}.
\newblock \emph{CoRR}, abs/2304.10547, 2023.
\newblock \doi{10.48550/arXiv.2304.10547}.
\newblock URL \url{https://arxiv.org/abs/2304.10547}.

\bibitem[{Mustafa Suleyman and Michael Bhaskar}(2023)]{comingWave}
{Mustafa Suleyman and Michael Bhaskar}.
\newblock \emph{The Coming Wave: Technology, Power, and the 21st Century's Greatest Dilemma}.
\newblock Crown, September 2023.

\bibitem[{OpenAI}(2018)]{openAICharter}
{OpenAI}.
\newblock {OpenAI Charter}, 2018.
\newblock URL \url{https://openai.com/charter}.
\newblock Accessed October 12, 2023.

\bibitem[{OpenAI}(2023)]{openAIMission}
{OpenAI}.
\newblock {OpenAI: About}, 2023.
\newblock URL \url{https://openai.com/about}.
\newblock Accessed October 12, 2023.

\bibitem[OpenAI(2023)]{openai2023gpt4}
OpenAI.
\newblock {GPT-4 Technical Report}.
\newblock \emph{CoRR}, abs/2303.08774, 2023.
\newblock \doi{10.48550/arXiv.2303.08774}.
\newblock URL \url{https://arxiv.org/abs/2303.08774}.

\bibitem[Papakyriakopoulos et~al.(2021)Papakyriakopoulos, Watkins, Winecoff, Jaźwińska, and Chattopadhyay]{papakyriakopoulos2021qualitative}
Papakyriakopoulos, O., Watkins, E.~A., Winecoff, A., Jaźwińska, K., and Chattopadhyay, T.
\newblock {Qualitative Analysis for Human Centered AI}.
\newblock \emph{CoRR}, abs/2112.03784, 2021.
\newblock \doi{10.48550/arXiv.2112.03784}.
\newblock URL \url{https://arxiv.org/abs/2112.03784}.

\bibitem[Parasuraman et~al.(2000)Parasuraman, Sheridan, and Wickens]{auto10levels}
Parasuraman, R., Sheridan, T., and Wickens, C.
\newblock A model for types and levels of human interaction with automation.
\newblock \emph{IEEE Transactions on Systems, Man, and Cybernetics - Part A: Systems and Humans}, 30\penalty0 (3):\penalty0 286--297, 2000.
\newblock \doi{10.1109/3468.844354}.

\bibitem[Pichai \& Hassabis(2023)Pichai and Hassabis]{geminiBlog}
Pichai, S. and Hassabis, D.
\newblock Introducing gemini: our largest and most capable ai model, December 2023.
\newblock URL \url{https://blog.google/technology/ai/google-gemini-ai/}.

\bibitem[Pressley et~al.(1987)Pressley, Borkowski, and Schneider]{pressleyMetacog}
Pressley, M., Borkowski, J., and Schneider, W.
\newblock Cognitive strategies: Good strategy users coordinate metacognition and knowledge.
\newblock \emph{Annals of Child Development}, 4:\penalty0 89--129, 1987.

\bibitem[PromptBase()]{promptbase}
PromptBase.
\newblock {PromptBase: Prompt Marketplace}.
\newblock URL \url{https://promptbase.com/}.
\newblock accessed on October 20, 2023.

\bibitem[Raji et~al.(2021)Raji, Bender, Paullada, Denton, and Hanna]{raji2021ai}
Raji, I.~D., Bender, E.~M., Paullada, A., Denton, E., and Hanna, A.
\newblock {AI and the Everything in the Whole Wide World Benchmark}.
\newblock \emph{CoRR}, abs/2111.15366, 2021.
\newblock \doi{10.48550/arXiv.2111.15366}.
\newblock URL \url{https://arxiv.org/abs/2111.15366}.

\bibitem[Ramesh et~al.(2022)Ramesh, Dhariwal, Nichol, Chu, and Chen]{dalle2}
Ramesh, A., Dhariwal, P., Nichol, A., Chu, C., and Chen, M.
\newblock {Hierarchical Text-Conditional Image Generation with CLIP Latents}.
\newblock April 2022.
\newblock URL \url{https://cdn.openai.com/papers/dall-e-2.pdf}.

\bibitem[R\"auker et~al.(2023)R\"auker, Ho, Casper, and Hadfield-Menell]{transparent}
R\"auker, T., Ho, A., Casper, S., and Hadfield-Menell, D.
\newblock {Toward Transparent AI: A Survey on Interpreting the Inner Structures of Deep Neural Networks}.
\newblock \emph{CoRR}, abs/2207.13243, 2023.
\newblock \doi{10.48550/arXiv.2207.13243}.
\newblock URL \url{https://arxiv.org/abs/2207.13243}.

\bibitem[Richmond \& McKinney(2009)Richmond and McKinney]{richmond2009biosafety}
Richmond, J.~Y. and McKinney, R.~W.
\newblock Biosafety in microbiological and biomedical laboratories, 2009.

\bibitem[Roy et~al.(2021)Roy, Posner, Barfoot, Beaudoin, Bengio, Bohg, Brock, Depatie, Fox, Koditschek, Lozano-Perez, Mansinghka, Pal, Richards, Sadigh, Schaal, Sukhatme, Therien, Toussaint, and de~Panne]{roy2021machine}
Roy, N., Posner, I., Barfoot, T., Beaudoin, P., Bengio, Y., Bohg, J., Brock, O., Depatie, I., Fox, D., Koditschek, D., Lozano-Perez, T., Mansinghka, V., Pal, C., Richards, B., Sadigh, D., Schaal, S., Sukhatme, G., Therien, D., Toussaint, M., and de~Panne, M.~V.
\newblock {From Machine Learning to Robotics: Challenges and Opportunities for Embodied Intelligence}.
\newblock \emph{CoRR}, abs/2110.15245, 2021.
\newblock \doi{10.48550/arXiv.2110.15245}.
\newblock URL \url{https://arxiv.org/abs/2110.15245}.

\bibitem[{SAE International}(2021)]{drivingLevels}
{SAE International}.
\newblock {Taxonomy and Definitions for Terms Related to Driving Automation Systems for On-Road Motor Vehicles}, April 2021.
\newblock URL \url{https://www.sae.org/standards/content/j3016_202104}.
\newblock Accessed October 12, 2023.

\bibitem[Saharia et~al.(2022)Saharia, Chan, Saxena, Li, Whang, Denton, Ghasemipour, Ayan, Mahdavi, Lopes, Salimans, Ho, Fleet, and Norouzi]{saharia2022photorealistic}
Saharia, C., Chan, W., Saxena, S., Li, L., Whang, J., Denton, E., Ghasemipour, S. K.~S., Ayan, B.~K., Mahdavi, S.~S., Lopes, R.~G., Salimans, T., Ho, J., Fleet, D.~J., and Norouzi, M.
\newblock {Photorealistic Text-to-Image Diffusion Models with Deep Language Understanding}.
\newblock \emph{CoRR}, abs/2205.11487, 2022.
\newblock \doi{10.48550/arXiv.2205.11487}.
\newblock URL \url{https://arxiv.org/abs/2205.11487}.

\bibitem[Schoenegger \& Park(2023)Schoenegger and Park]{schoenegger2023large}
Schoenegger, P. and Park, P.~S.
\newblock Large language model prediction capabilities: Evidence from a real-world forecasting tournament, 2023.

\bibitem[Searle(1980)]{searle_1980}
Searle, J.~R.
\newblock {Minds, Brains, and Programs}.
\newblock \emph{Behavioral and Brain Sciences}, 3:\penalty0 417–424, 1980.
\newblock \doi{10.1017/S0140525X00005756}.

\bibitem[Serapio-García et~al.(2023)Serapio-García, Safdari, Crepy, Sun, Fitz, Romero, Abdulhai, Faust, and Matarić]{serapio2023personality}
Serapio-García, G., Safdari, M., Crepy, C., Sun, L., Fitz, S., Romero, P., Abdulhai, M., Faust, A., and Matarić, M.
\newblock {Personality Traits in Large Language Models}.
\newblock \emph{CoRR}, abs/2307.00184, 2023.
\newblock \doi{10.48550/arXiv.2307.00184}.
\newblock URL \url{https://arxiv.org/abs/2307.00184}.

\bibitem[Shah et~al.(2021)Shah, Freire, Alex, Freedman, Krasheninnikov, Chan, Dennis, Abbeel, Dragan, and Russell]{shah2021benefits}
Shah, R., Freire, P., Alex, N., Freedman, R., Krasheninnikov, D., Chan, L., Dennis, M.~D., Abbeel, P., Dragan, A., and Russell, S.
\newblock {Benefits of Assistance over Reward Learning}, 2021.
\newblock URL \url{https://openreview.net/forum?id=DFIoGDZejIB}.

\bibitem[Shah et~al.(2025)Shah, Irpan, Turner, Wang, Conmy, Lindner, Brown-Cohen, Ho, Nanda, Popa, Jain, Greig, Albanie, Emmons, Farquhar, Krier, Rajamanoharan, Bridgers, Ijitoye, Everitt, Krakovna, Varma, Mikulik, Kenton, Orr, Legg, Goodman, Dafoe, Flynn, and Dragan]{shah2025approachtechnicalagisafety}
Shah, R., Irpan, A., Turner, A.~M., Wang, A., Conmy, A., Lindner, D., Brown-Cohen, J., Ho, L., Nanda, N., Popa, R.~A., Jain, R., Greig, R., Albanie, S., Emmons, S., Farquhar, S., Krier, S., Rajamanoharan, S., Bridgers, S., Ijitoye, T., Everitt, T., Krakovna, V., Varma, V., Mikulik, V., Kenton, Z., Orr, D., Legg, S., Goodman, N., Dafoe, A., Flynn, F., and Dragan, A.
\newblock An approach to technical agi safety and security, 2025.
\newblock URL \url{https://arxiv.org/abs/2504.01849}.

\bibitem[Shanahan(2010)]{murrayEmbody}
Shanahan, M.
\newblock \emph{{Embodiment and the Inner Life}}.
\newblock Oxford University Press, 2010.

\bibitem[Shanahan(2015)]{murrayBook}
Shanahan, M.
\newblock \emph{{The Technological Singularity}}.
\newblock MIT Press, August 2015.

\bibitem[Sheridan \& Parasuraman(2005)Sheridan and Parasuraman]{auto8levels}
Sheridan, T.~B. and Parasuraman, R.
\newblock Human-automation interaction.
\newblock \emph{Reviews of Human Factors and Ergonomics}, 1\penalty0 (1):\penalty0 89--129, 2005.
\newblock \doi{10.1518/155723405783703082}.
\newblock URL \url{https://doi.org/10.1518/155723405783703082}.

\bibitem[Sheridan et~al.(1978)Sheridan, Verplank, and Brooks]{sheridan1978human}
Sheridan, T.~B., Verplank, W.~L., and Brooks, T.
\newblock Human/computer control of undersea teleoperators.
\newblock In \emph{NASA. Ames Res. Center The 14th Ann. Conf. on Manual Control}, 1978.

\bibitem[Shevlane et~al.(2023)Shevlane, Farquhar, Garfinkel, Phuong, Whittlestone, Leung, Kokotajlo, Marchal, Anderljung, Kolt, Ho, Siddarth, Avin, Hawkins, Kim, Gabriel, Bolina, Clark, Bengio, Christiano, and Dafoe]{shevlane2023model}
Shevlane, T., Farquhar, S., Garfinkel, B., Phuong, M., Whittlestone, J., Leung, J., Kokotajlo, D., Marchal, N., Anderljung, M., Kolt, N., Ho, L., Siddarth, D., Avin, S., Hawkins, W., Kim, B., Gabriel, I., Bolina, V., Clark, J., Bengio, Y., Christiano, P., and Dafoe, A.
\newblock {Model evaluation for extreme risks}.
\newblock \emph{CoRR}, abs/2305.15324, 2023.
\newblock \doi{10.48550/arXiv.2305.15324}.
\newblock URL \url{https://arxiv.org/abs/2305.15324}.

\bibitem[Shneiderman(2020)]{shneiderman2020humancentered}
Shneiderman, B.
\newblock Human-centered artificial intelligence: Reliable, safe \& trustworthy, 2020.
\newblock URL \url{https://arxiv.org/abs/2002.04087v1}.

\bibitem[Silver et~al.(2016)Silver, Huang, Maddison, Guez, Sifre, van~den Driessche, Schrittwieser, Antonoglou, Panneershelvam, Lanctot, Dieleman, Grewe, Nham, Kalchbrenner, Sutskever, Lillicrap, Leach, Kavukcuoglu, Graepel, and Hassabis]{alphago}
Silver, D., Huang, A., Maddison, C.~J., Guez, A., Sifre, L., van~den Driessche, G., Schrittwieser, J., Antonoglou, I., Panneershelvam, V., Lanctot, M., Dieleman, S., Grewe, D., Nham, J., Kalchbrenner, N., Sutskever, I., Lillicrap, T., Leach, M., Kavukcuoglu, K., Graepel, T., and Hassabis, D.
\newblock {Mastering the Game of Go with Deep Neural Networks and Tree Search}.
\newblock \emph{Nature}, 529:\penalty0 484--489, 2016.
\newblock \doi{10.1038/nature16961}.

\bibitem[Silver et~al.(2017)Silver, Schrittwieser, Simonyan, Antonoglou, Huang, Guez, Hubert, Baker, Lai, Bolton, Chen, Lillicrap, Hui, Sifre, van~den Driessche, Graepel, and Hassabis]{alphagoRL}
Silver, D., Schrittwieser, J., Simonyan, K., Antonoglou, I., Huang, A., Guez, A., Hubert, T., Baker, L., Lai, M., Bolton, A., Chen, Y., Lillicrap, T., Hui, F., Sifre, L., van~den Driessche, G., Graepel, T., and Hassabis, D.
\newblock {Mastering the Game of Go Without Human Knowledge}.
\newblock \emph{Nature}, 550:\penalty0 354--359, 2017.
\newblock \doi{10.1038/nature24270}.

\bibitem[Silver et~al.(2018)Silver, Hubert, Schrittwieser, Antonoglou, Lai, Guez, Lanctot, Sifre, Kumaran, Graepel, Lillicrap, Simonyan, and Hassabis]{alphazero}
Silver, D., Hubert, T., Schrittwieser, J., Antonoglou, I., Lai, M., Guez, A., Lanctot, M., Sifre, L., Kumaran, D., Graepel, T., Lillicrap, T., Simonyan, K., and Hassabis, D.
\newblock {A General Reinforcement Learning Algorithm that Masters Chess, Shogi, and Go through Self-play}.
\newblock \emph{Science}, 362\penalty0 (6419):\penalty0 1140--1144, 2018.
\newblock \doi{10.1126/science.aar6404}.
\newblock URL \url{https://www.science.org/doi/abs/10.1126/science.aar6404}.

\bibitem[Srivastava et~al.(2023)Srivastava, Rastogi, Rao, and et~al.]{srivastava2023imitation}
Srivastava, A., Rastogi, A., Rao, A., and et~al.
\newblock {Beyond the Imitation Game: Quantifying and Extrapolating the Capabilities of Language Models}.
\newblock \emph{CoRR}, abs/2206.04615, 2023.
\newblock \doi{10.48550/arXiv.2206.04615}.
\newblock URL \url{https://arxiv.org/abs/2206.04615}.

\bibitem[Stockfish(2023)]{stockfish}
Stockfish.
\newblock {Stockfish - Open Source Chess Engine}, 2023.
\newblock URL \url{https://stockfishchess.org/}.

\bibitem[Tang et~al.(2023)Tang, LeBel, Jain, and Huth]{neuralUI}
Tang, J., LeBel, A., Jain, S., and Huth, A.~G.
\newblock {Semantic Reconstruction of Continuous Language from Non-invasive Brain Recordings}.
\newblock \emph{Nature Neuroscience}, 26:\penalty0 858--866, 2023.
\newblock \doi{10.1038/s41593-023-01304-9}.

\bibitem[Terry et~al.(2023)Terry, Kulkarni, Wattenberg, Dixon, and Morris]{terryHCIAlignment}
Terry, M., Kulkarni, C., Wattenberg, M., Dixon, L., and Morris, M.~R.
\newblock {AI Alignment in the Design of Interactive AI: Specification Alignment, Process Alignment, and Evaluation Support}.
\newblock \emph{CoRR}, abs/2311.00710, 2023.
\newblock \doi{10.48550/arXiv.2311.00710}.
\newblock URL \url{https://arxiv.org/abs/2311.00710}.

\bibitem[Touvron et~al.(2023)Touvron, Martin, Stone, Albert, Almahairi, Babaei, Bashlykov, Batra, Bhargava, Bhosale, Bikel, Blecher, Ferrer, Chen, Cucurull, Esiobu, Fernandes, Fu, Fu, Fuller, Gao, Goswami, Goyal, Hartshorn, Hosseini, Hou, Inan, Kardas, Kerkez, Khabsa, Kloumann, Korenev, Koura, Lachaux, Lavril, Lee, Liskovich, Lu, Mao, Martinet, Mihaylov, Mishra, Molybog, Nie, Poulton, Reizenstein, Rungta, Saladi, Schelten, Silva, Smith, Subramanian, Tan, Tang, Taylor, Williams, Kuan, Xu, Yan, Zarov, Zhang, Fan, Kambadur, Narang, Rodriguez, Stojnic, Edunov, and Scialom]{touvron2023llama}
Touvron, H., Martin, L., Stone, K., Albert, P., Almahairi, A., Babaei, Y., Bashlykov, N., Batra, S., Bhargava, P., Bhosale, S., Bikel, D., Blecher, L., Ferrer, C.~C., Chen, M., Cucurull, G., Esiobu, D., Fernandes, J., Fu, J., Fu, W., Fuller, B., Gao, C., Goswami, V., Goyal, N., Hartshorn, A., Hosseini, S., Hou, R., Inan, H., Kardas, M., Kerkez, V., Khabsa, M., Kloumann, I., Korenev, A., Koura, P.~S., Lachaux, M.-A., Lavril, T., Lee, J., Liskovich, D., Lu, Y., Mao, Y., Martinet, X., Mihaylov, T., Mishra, P., Molybog, I., Nie, Y., Poulton, A., Reizenstein, J., Rungta, R., Saladi, K., Schelten, A., Silva, R., Smith, E.~M., Subramanian, R., Tan, X.~E., Tang, B., Taylor, R., Williams, A., Kuan, J.~X., Xu, P., Yan, Z., Zarov, I., Zhang, Y., Fan, A., Kambadur, M., Narang, S., Rodriguez, A., Stojnic, R., Edunov, S., and Scialom, T.
\newblock {Llama 2: Open Foundation and Fine-Tuned Chat Models}, 2023.

\bibitem[Tullis \& Fraundorf(2017)Tullis and Fraundorf]{TullisMetacog}
Tullis, J. and Fraundorf, S.
\newblock Predicting others’ memory performance: The accuracy and bases of social metacognition.
\newblock \emph{Journal of Memory and Language}, 95:\penalty0 124--137, 2017.
\newblock \doi{https://doi.org/10.1016/j.jml.2017.03.003}.

\bibitem[Turing(1950)]{turingTest}
Turing, A.
\newblock {Computing Machinery and Intelligence}.
\newblock \emph{Mind}, LIX:\penalty0 433--460, October 1950.
\newblock URL \url{https://doi.org/10.1093/mind/LIX.236.433}.

\bibitem[Varadi et~al.(2021)Varadi, Anyango, Deshpande, Nair, Natassia, Yordanova, Yuan, Stroe, Wood, Laydon, Žídek, Green, Tunyasuvunakool, Petersen, Jumper, Clancy, Green, Vora, Lutfi, Figurnov, Cowie, Hobbs, Kohli, Kleywegt, Birney, Hassabis, and Velankar]{alphafold2}
Varadi, M., Anyango, S., Deshpande, M., Nair, S., Natassia, C., Yordanova, G., Yuan, D., Stroe, O., Wood, G., Laydon, A., Žídek, A., Green, T., Tunyasuvunakool, K., Petersen, S., Jumper, J., Clancy, E., Green, R., Vora, A., Lutfi, M., Figurnov, M., Cowie, A., Hobbs, N., Kohli, P., Kleywegt, G., Birney, E., Hassabis, D., and Velankar, S.
\newblock {AlphaFold Protein Structure Database: Massively Expanding the Structural Coverage of Protein-Sequence Space with High-Accuracy Models}.
\newblock \emph{Nucleic Acids Research}, 50:\penalty0 D439--D444, 11 2021.
\newblock ISSN 0305-1048.
\newblock \doi{10.1093/nar/gkab1061}.
\newblock URL \url{https://doi.org/10.1093/nar/gkab1061}.

\bibitem[Vaswani et~al.(2023)Vaswani, Shazeer, Parmar, Uszkoreit, Jones, Gomez, Kaiser, and Polosukhin]{vaswani2023attention}
Vaswani, A., Shazeer, N., Parmar, N., Uszkoreit, J., Jones, L., Gomez, A.~N., Kaiser, L., and Polosukhin, I.
\newblock {Attention Is All You Need}.
\newblock \emph{CoRR}, abs/1706.03762, 2023.
\newblock \doi{10.48550/arXiv.1706.03762}.
\newblock URL \url{https://arxiv.org/abs/1706.03762}.

\bibitem[Veerabadran et~al.(2023)Veerabadran, Goldman, Shankar, and et~al.]{subliminal}
Veerabadran, V., Goldman, J., Shankar, S., and et~al.
\newblock {Subtle Adversarial Image Manipulations Influence Both Human and Machine Perception}.
\newblock \emph{Nature Communications}, 14, 2023.
\newblock \doi{10.1038/s41467-023-40499-0}.

\bibitem[Webb et~al.(2023)Webb, Holyoak, and Lu]{logic}
Webb, T., Holyoak, K.~J., and Lu, H.
\newblock {Emergent Analogical Reasoning in Large Language Models}.
\newblock \emph{Nature Human Behavior}, 7:\penalty0 1526–1541, 2023.
\newblock URL \url{https://doi.org/10.1038/s41562-023-01659-w}.

\bibitem[Wei et~al.(2022)Wei, Tay, Bommasani, Raffel, Zoph, Borgeaud, Yogatama, Bosma, Zhou, Metzler, Chi, Hashimoto, Vinyals, Liang, Dean, and Fedus]{wei2022emergent}
Wei, J., Tay, Y., Bommasani, R., Raffel, C., Zoph, B., Borgeaud, S., Yogatama, D., Bosma, M., Zhou, D., Metzler, D., Chi, E.~H., Hashimoto, T., Vinyals, O., Liang, P., Dean, J., and Fedus, W.
\newblock {Emergent Abilities of Large Language Models}.
\newblock \emph{CoRR}, abs/2206.07682, 2022.
\newblock \doi{10.48550/arXiv.2206.07682}.
\newblock URL \url{https://arxiv.org/abs/2206.07682}.

\bibitem[Weizenbaum(1966)]{eliza}
Weizenbaum, J.
\newblock {ELIZA—a Computer Program for the Study of Natural Language Communication between Man and Machine}.
\newblock \emph{Commun. ACM}, 9\penalty0 (1):\penalty0 36–45, jan 1966.
\newblock ISSN 0001-0782.
\newblock \doi{10.1145/365153.365168}.
\newblock URL \url{https://doi.org/10.1145/365153.365168}.

\bibitem[Wiggers(2021)]{roboticsCut}
Wiggers, K.
\newblock {OpenAI Disbands its Robotics Research Team}.
\newblock VentureBeat, July 2021.

\bibitem[Wikipedia(2023{\natexlab{a}})]{goostman}
Wikipedia.
\newblock {Eugene Goostman - Wikipedia, The Free Encyclopedia}.
\newblock https://en.wikipedia.org/wiki/Eugene\_Goostman, 2023{\natexlab{a}}.
\newblock Accessed October 12, 2023.

\bibitem[Wikipedia(2023{\natexlab{b}})]{turingWeaknesses}
Wikipedia.
\newblock {Turing Test: Weaknesses --- Wikipedia, The Free Encyclopedia}.
\newblock https://en.wikipedia.org/wiki/Turing\_test, 2023{\natexlab{b}}.
\newblock Accessed October 12, 2023.

\bibitem[Winograd(1971)]{shrdlu}
Winograd, T.
\newblock {Procedures as a Representation for Data in a Computer Program for Understanding Natural Language}.
\newblock \emph{MIT AI Technical Reports}, 1971.

\bibitem[Wozniak(2010)]{wozVideo}
Wozniak, S.
\newblock {Could a Computer Make a Cup of Coffee?}
\newblock Fast Company interview: https://www.youtube.com/watch?v=MowergwQR5Y, 2010.

\bibitem[Yang et~al.(2023)Yang, Li, Lin, Wang, Lin, Liu, and Wang]{yang2023dawn}
Yang, Z., Li, L., Lin, K., Wang, J., Lin, C.-C., Liu, Z., and Wang, L.
\newblock {The Dawn of LMMs: Preliminary Explorations with GPT-4V(ision)}.
\newblock \emph{CoRR}, abs/2309.17421, 2023.
\newblock \doi{10.48550/arXiv.2309.17421}.
\newblock URL \url{https://arxiv.org/abs/2309.17421}.

\bibitem[Zamfirescu-Pereira et~al.(2023)Zamfirescu-Pereira, Wong, Hartmann, and Yang]{cantprompt}
Zamfirescu-Pereira, J., Wong, R.~Y., Hartmann, B., and Yang, Q.
\newblock Why johnny can’t prompt: How non-ai experts try (and fail) to design llm prompts.
\newblock In \emph{Proceedings of the 2023 CHI Conference on Human Factors in Computing Systems}, CHI '23, New York, NY, USA, 2023. Association for Computing Machinery.
\newblock ISBN 9781450394215.
\newblock \doi{10.1145/3544548.3581388}.
\newblock URL \url{https://doi.org/10.1145/3544548.3581388}.

\bibitem[Zwetsloot \& Dafoe(2019)Zwetsloot and Dafoe]{zwetsloot2019thinking}
Zwetsloot, R. and Dafoe, A.
\newblock {Thinking about Risks from AI: Accidents, Misuse and Structure}.
\newblock \emph{Lawfare}, 11:\penalty0 2019, 2019.

\end{thebibliography}
\bibliographystyle{icml2023}





\end{document}